\documentclass[a4paper,conference]{IEEEtran}

\usepackage{epsfig}
\usepackage{graphicx}
\usepackage{amsmath}
\usepackage{amssymb}
\usepackage{caption}
\usepackage{hyperref}

\def\ie{\emph{i.e}.}

\begin{document}

\pagebreak
\pagebreak

%
\title{Aggregating Layers for Deepfake Detection}

\author{Amir Jevnisek\\
School of Electrical Engineering\\
Tel-Aviv University\\
Tel-Aviv, Israel\\
{\tt\small amirjevn@mail.tau.ac.il}
\and
Shai Avidan\\
School of Electrical Engineering\\
Tel-Aviv University\\
Tel-Aviv, Israel\\
{\tt\small avidan@eng.tau.ac.il}
}

\maketitle

\begin{abstract}
  The increasing popularity of facial manipulation (Deepfakes) and synthetic face creation raises the need to develop robust forgery detection solutions. Crucially, most work in this domain assume that the Deepfakes in the test set come from the same Deepfake algorithms that were used for training the network. This is not how things work in practice. Instead, we consider the case where the network is trained on one Deepfake algorithm, and tested on Deepfakes generated by another algorithm. Typically, supervised techniques follow a pipeline of visual feature extraction from a deep backbone, followed by a binary classification head. Instead, our algorithm aggregates features extracted across all layers of one backbone network to detect a fake. We evaluate our approach on two domains of interest - Deepfake detection and Synthetic image detection, and find that we achieve SOTA results. 
  

\end{abstract}

\section{Introduction}
High quality facial manipulations are no longer within the purview of the research community. Facial Forgery (Deepfakes) tools are widely spread and available to all. The Reface App \cite{ReFaceApp}, for example, replaces ones face with Captain Jack Sparrow's \footnote {Fictional character from the Pirates of the Caribbean movie Series.} face from a few images of the target face and within a couple of seconds. While this is an example of an entertaining use of facial manipulations, the use of this technology might pose threats to privacy, democracy and national security \cite{Chesney2018DeepFA}. Therefore, it is clear that forgery detection algorithms are needed.

It is common \cite{Tolosana2020DeepFakesAB} to categorize facial manipulations to four families of manipulations: i) entire face synthesis, ii) identity swap, iii) attribute manipulation, and iv) expression swap which we will address as facial reenactment. 
Entire face synthesis is a category in which random noise is served as an input to a system and a fully synthesized face image is generated. Identity Swap, is a family of methods in which a source face is blended into a target face image. The outcome is a blend of the targets’ context and the sources’ identity. Attribute manipulation takes one facial attribute such as “wears eyeglasses” or “hair color” and changes that attribute. Facial Reenactment, on the other hand, preserves both the context and the identity but replaces the gestures made by an “actor” (source) video with those of the target.

Most research in the field assumes that the training set and test set come from the same distribution. That is, a collection of Deepfake images, created by a number of Deepfake algorithms, is randomly split into train and test set and the goal of the Deepfake detector is to correctly distinguish fake images from real.

We argue that this is {\em not} how a deepfake detector will be used in practice. In practice, the detector will be trained on Deepfake images produced by one algorithm and will have to detect Deepfake images produced by a yet to be developed and unknown Deepfake algorithm. This is the setting of this paper.

A straight-forward approach to Deepfake detection is to rely on some backbone neural network (i.e., resnet, VGG, EfficientNet) with a binary classification head. This assumes that data propagates through the network until it reaches the classification head that determines if the image is real or fake.

We improve the performance of the backbone network by aggregating information from all layers of the network. Specifically, we use skip connections from every layer of the network to the fully-connected classification head. This way, various features, that correspond to different receptive field in the image plane, are all used, at once, by the classification head.

Comparing the performance of Deepfake detection is usually done using the Average Precision metric. However, when the detector is trained on a set coming from one Deepfake algorithm and tested on a set coming from a different Deepfake algorithm, we need a way to rank the competing algorithms. To this end, we suggest using a popular measure, the Coefficient of Variation (CoV) of Average Precision scores, to measure the performance of the various algorithms on different datasets. We use these measures to report results on both synthetic image detection, as well as Deepfake detection on standarad datasets. In summary, our contributions are threefold:
\begin{itemize}
  \item A new architecture to fine-tune a backbone network, using its layers.
  \item Works on both Deepfake and Synthetic Image detection.
  \item SOTA overall performance for cross dataset generalization (training on one Deepfake or synthetic image model, and testing on another).
\end{itemize}

\section{Related Works}
\subsection{Forgery Detection Techniques}
Forgery detection techniques can be roughly divided into three categories. The first class is based on Spatial/Frequency methods. These methods are based on some well-engineered cues that are extracted from the image. Such cues have been thoughtfully investigated in \cite{patchforensics} while more face-specific ({i.e.\ } physiological) cues are discussed in \cite{hu2020exposing}. To this category we can also attribute multi-task methods that attempt to find CNN artifacts \cite{Zhou2017TwoStreamNN} in images or inconsistency in non-localized features \cite{Zhao2020LearningTR}.

In the frequency domain, \cite{li2021frequency} use a DCT-coefficient rearrangement block and a feature extraction block to mine frequency cues in a data-driven approach. On top of that, their main contribution is a loss which promotes intra-class compactness. It encourages pristine images to have a minimal distance to some center point, and a larger distance for manipulated faces by at least a margin. \cite{wang2021representative} takes the route of data augmentation to guide a detector to refine and enlarge its attention. They compute the Top-$N$ sensitive facial regions using a gradient-based method. They then occlude these regions with random integers, and the resulting image is served back to the model with the same label. This process allows the model to mine for features it ignored before. 

The second category exploits temporal features. \cite{Montserrat2020DeepfakesDW} extract visual and temporal features using both CNN and RNN. CNNs are used to extract visual cues from each frame and RNNs are used to aggregate features from all regions in all frames. \cite{sun2021improving} and \cite{haliassos2021lips} use facial landmarks extracted from a sequence of frames to distinguish real from fake videos. It follow that these methods measure their performance at a video-level. 

\cite{sun2021improving} use the facial landmark locations and velocity as inputs to two RNNs which are, in turn, aggregated to a final prediction score. \cite{haliassos2021lips} use a pretrained backbone on an auxiliary task: visual speech recognition (lipreading). The features extracted from every frame of the video are then aggregated to a temporal network that produces the final real/fake classification verdict.

The third category is based on anomaly detection. The goal of such methods is to train solely on pristine images and output a normality score of the query image. One can think of the normality score as a measure of how real the query image is or, put another way, the score can be thought as the inverse of an “out-of-distribution” score. Kahlid {\emph{et al.}}~ \cite{Khalid2020OCFakeDectCD} use both reconstruction loss and latent space distance to predict real from fake images. \cite{Aneja2020GeneralizedZA} map the real and fake classes to Gaussian distributions and measure the distribution alignment distance.

\subsection{Manipulations}
Image manipulations are at the core of computer vision tasks. They range from image enhancement to image splicing and blending which can be applied to any natural image. Facial image manipulations for identity swapping are closely related to the aforementioned tasks. Deepfakes~\cite{DeepfakesGithub} and FaceSwap~\cite{FaceSwapGithub} are examples of identity swapping techniques based on deep learning techniques. On the other hand, Face2Face~\cite{Thies2019Face2FaceRF} and NeuralTextures~\cite{Thies2019DeferredNR} are examples of Facial Reenactment techniques. Some datasets do not state the manipulation technique (or techniques) used to generate the manipulations. 

\subsection{Deepfake Datasets}
\textbf{FaceForensics++} \cite{roessler2019faceforensicspp} is a common benchmark for fake detection of human faces. This dataset consists of four manipulations: Deepfakes~\cite{DeepfakesGithub}, Face2Face~\cite{Thies2019Face2FaceRF}, FaceSwap~\cite{FaceSwapGithub} and NeuralTextures~\cite{Thies2019DeferredNR} applied to a set of $1,000$ pristine videos. The videos are available in three compression modes: Raw (c0), High Quality (c23) and Low Quality (c40).

\begin{figure}[t]
\begin{center}
\includegraphics[width=1.0\linewidth]{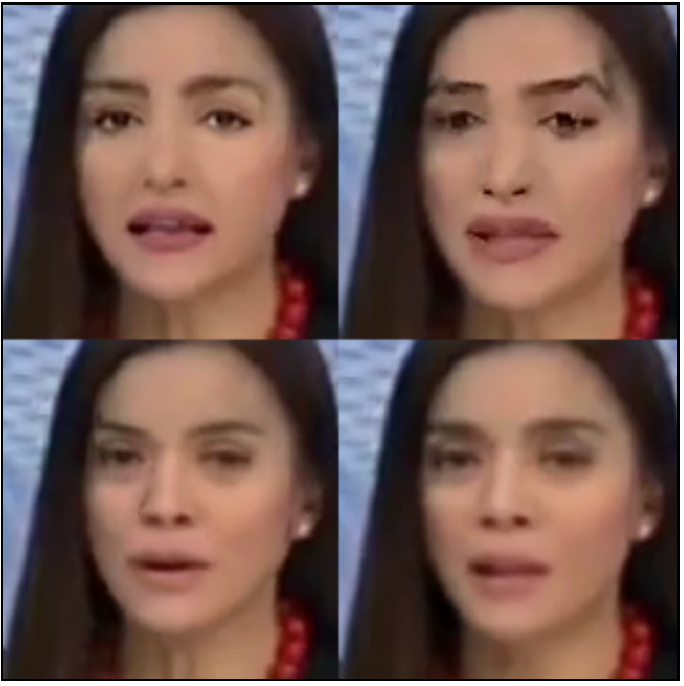}
\end{center}
  \caption{\textbf{Deepfakes} FaceForensics++ fake images examples. Top to bottom and left to right: Deepfakes, FaceSwap, Face2Face and NeuralTextures.}
\label{fig:DeepfakeExamples}
\end{figure}

\subsection{Synthetic Images}
We consider synthetic facial images created by fully generative models. We use three types of such generative methods: Generative Adversarial methods, Flow-based generative methods and Gaussian Mixture Models (GMMs). Generative adversarial methods map noise from some distribution to an image that looks like a face. The building blocks of these methods are a generative component, that is responsible for the image creation, and a discriminative component, that helps the generator to train by distinguishing pristine from synthetic images. Datasets of synthetic faces differ in the method used to create synthetic images and the dataset of original images that they were adversarially trained against. 

\textbf{Progressive-GAN (PGAN)}~\cite{karras2017progressive} is a method which is trained gradually, to first build low resolution images of synthetic faces and as the training evolves, add new layers to the generator and discriminator which in turn adds fine details to the result images.

\textbf{StyleGAN}~\cite{karras2018style} is a method which is essentially an alternative generator architecture for generative adversarial networks. While traditional generators feed the latent representations into the input layer, they first map the input to an intermediate latent space for which each feature is responsible for a different high-level attribute in the target image.

We follow~\cite{chai2020makes} and use PGAN which was trained on CelebA-HQ, StyleGAN trained on CelebA-HQ and Flickr-Faces-HQ (FFHQ) dataset, StyleGAN2 on FFHQ dataset.

The flow based algorithm that we use for synthetic image creation is \textbf{Glow}~\cite{kingma2018glow}. This method is a simple type of generative flow using an invertible $1 \times 1$ convolution. Contrary to Generative Adversarial Networks, this model is optimized on a plain log-likelihood objective.

We take~\cite{richardson2018gans} as a face generator based on Gaussian Mixture Models. The advantage of GMM synthetic images over GAN synthesized images is that GMM synthetic images generate samples which capture the full distribution of natural images. We follow~\cite{chai2020makes} and use GMM trained on CelebA~\cite{liu2015deep}, and Glow trained on CelebA-HQ.

\begin{figure}[t]
\begin{center}
\includegraphics[width=1.0\linewidth]{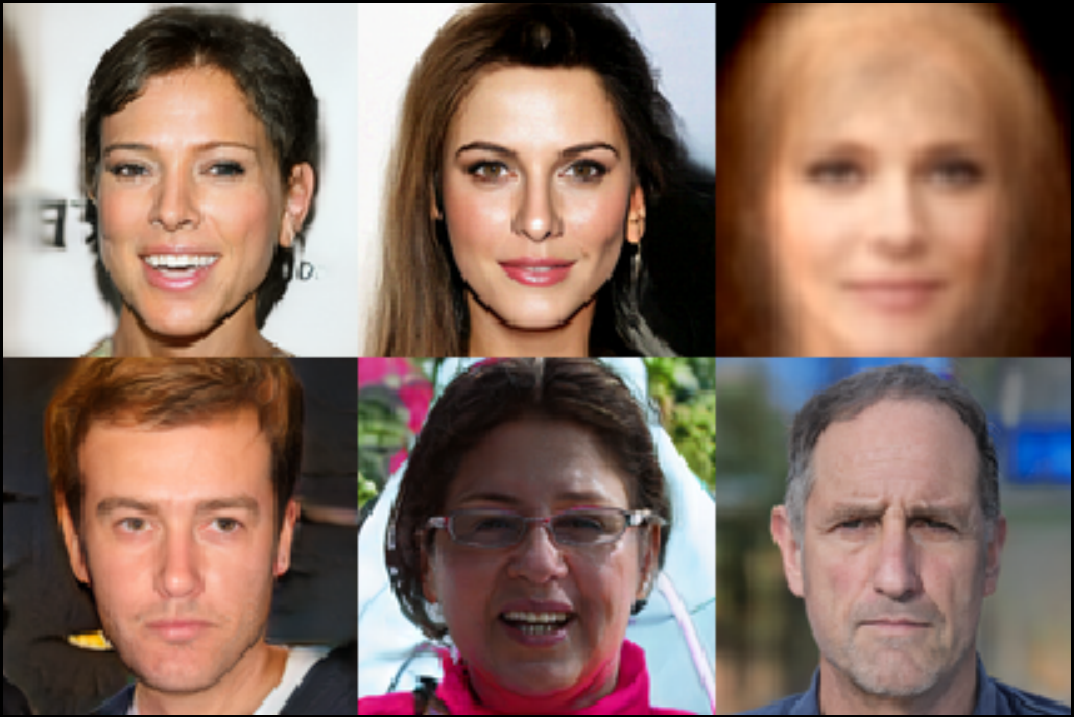}
\end{center}
  \caption{\textbf{Synthetic Images} GAN generated images, top to bottom, right to left: SGAN, Glow and GMM trained with CelebA-HQ, PGAN, SGAN and SGAN2 trained with FFHQ.}
\label{fig:SyntheticImagesExamples}
\end{figure}


\subsection{Finding features}
In \cite{chai2020makes} the authors wish to understand what makes fake images detectable. To do that, they generate a classifier for each and every single patch in the image. Each patch classifier gives two scores for how real and fake the image is. The scores are then summed over all image patches and the highest score is the prediction for the entire image. This allows~\cite{chai2020makes} to draw heatmaps based on the patch-wise predictions, and from that deduce which parts of the image are predominant in the classification. Moreover, they segment facial images to semantically meaningful partitions such as hair, eyes, nose, mouth, ear and background. They generate the distribution of the most predictive patch in each image and associate each of these patches to the semantically meaningful partition. Using this procedure they can associate the most predictive patch to a semantically meaningful region. Their strongest cues were found around the hair and the background.

We, on the other hand, take the conjugate direction. Instead of a spatial answer as to what makes fake images detectable, we answer this question in the features domain. Our network aggregates all features of a single backbone and can be used for both detection and analysis of which features contribute to the classification. The detection is trivial, an inference through our model yields a score, which is a measure of how fake the image is. Our model can also be used for feature analysis. The inference yields primitives for every feature and essentially performs linear regression over the concatenation of all the primitives. The product of the weight and primitive is a proxy of the importance of each primitive to the final classification score. A product of primitives and weights gives an importance score for each feature. We provide this analysis in section \ref{Experiments}.

\subsection{Deep Layer Aggregation}
Deep Layer Aggregation~\cite{yu2018deep} define an aggregation to be a combination of different layers through the network, and call it \textit{deep} if the earliest aggregated layer passes through multiple aggregations. They propose two types of aggregation mechanisms - iterative aggregation, which is used to fuse resolutions and scales, and hierarchical aggregation, which merges features from all network modules and channels. Finally, they propose an architecture that merges both aggregation mechanisms - Deep Layer Aggregation. This approach breaks the basic architecture of the network, due to the hierarchical layer aggregation mechanism that involves routing intermediate aggregations back into the network.

We, on the other hand, consider shallow aggregation of the backbone layers with concatenation of the processed features and a final aggregation block which is a simple linear regressor. This approach has several advantages. First, it preserves the backbone architecture structure in the forward pass of the network. Second, it lets the network architect understand which features are important to make the classification. Finally, it allows skip connections for the gradients to flow from at least two sources i.e., a direct route from the final prediction score and a gradient from the successive feature extractor.

\section{Method}

\begin{figure*}[t]
\begin{center}
\begin{tabular}{ccc}
\includegraphics[height=0.25\textheight]{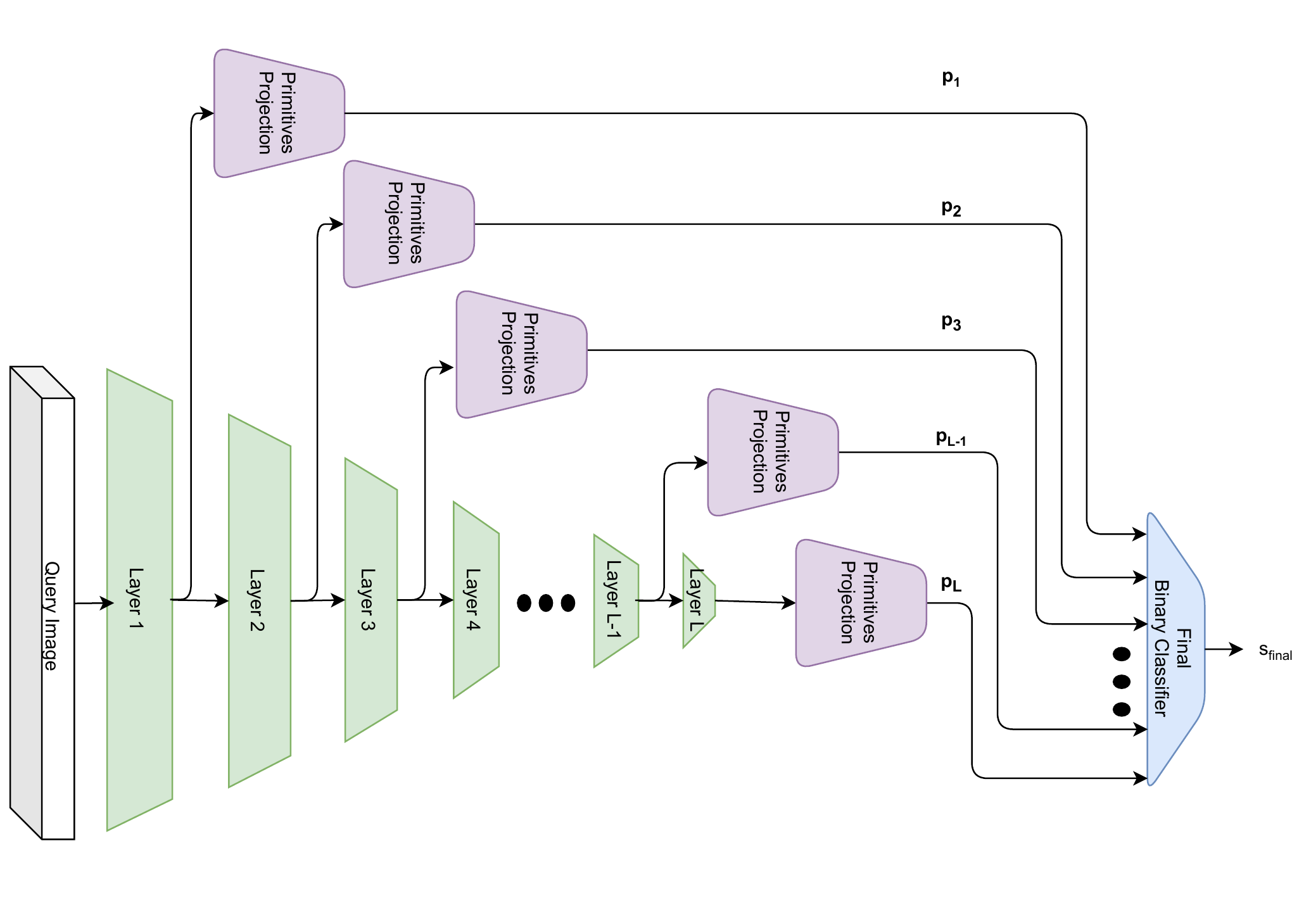} & &
\includegraphics[height=0.25\textheight]{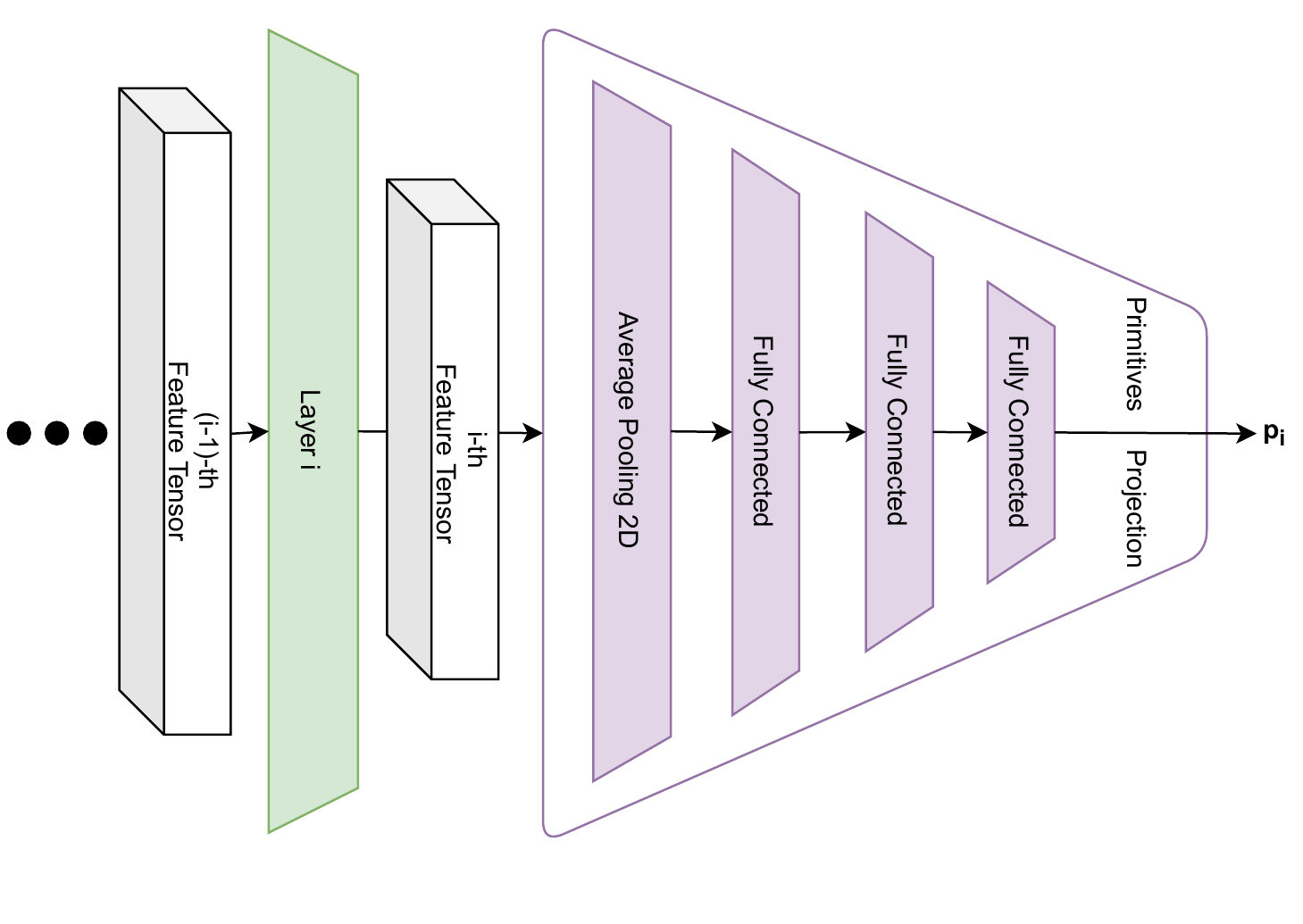} \\
(Overall model architecture) & & (Primitive extraction block) \\
\end{tabular}
\end{center}
  \caption{{\bf Architecture: } (Left) Overall model architecture. We project each feature tensor the backbone computes to a set of primitives. Primitives from features in all scales and receptive fields are aggregated using a simple Linear Regression block to produce the final classification score $s_{final}$. (Right) Primitives extraction block. For each layer, we project the feature tensor at its output to a small vector $p_i$ of $10$ primitives. }
\label{fig:long}
\label{fig:ModelArchitecture}
\end{figure*}

Our method is based on a simple and direct aggregation of all features extracted from a backbone to a Linear Regression model which outputs a score for how unreal the image is. We constructed the architecture to be backbone agnostic. Features extracted from all layers / blocks or endpoints of the backbone are summarized to a set of numbers we term primitives. Primitives from all features are aggregated with a linear regressor to a final score. Thus, there exists a direct link between each feature summary and the final classification score. This allows us to enjoy a performance gain w.r.t existing models which are designed to be interpretable. Moreover, with a trained model of our architecture we can easily spot which features contribute the most to the ability to detect fake images.

\begin{table*}[h]

\begin{center}
\begin{tabular}{|l|c|c|c|c|c|c|c|c|c|c|}

\hline
\multicolumn{1}{|c|}{}& \multicolumn{4}{|c|}{\textbf{CelebA-HQ dataset}} & \multicolumn{3}{|c|}{\textbf{FFHQ dataset}} & \multicolumn{3}{|c|}{\textbf{CoV}} \\

\textbf{Model } & \textbf{PGAN} & \textbf{SGAN} & \textbf{Glow} & \textbf{GMM} & \textbf{PGAN} & \textbf{SGAN} & \textbf{SGAN2} &  \textbf{mean} &  \textbf{std} & \textbf{$CoV^{-1}$} \\
\hline
\hline
Resnet Layer 1 & 100.0 & 97.22 & 72.80 & 80.69 & 99.81 & 72.91 & 71.81 & 82.54 & 11.69 & 7.06 \\
Xception Block 1 & 100.0 & 98.68 & 95.48 & 76.21 & 99.68 & 81.35 & 77.40 & 88.13 & 10.02 & 8.80 \\
Xception Block 2 & 100.0 & 99.99 & 67.49 & \textbf{91.38} & \textbf{100.0} & 90.12 & 90.85 & 89.97 & 10.87 & 8.28 \\
Xception Block 3 & 100.0 & \textbf{100.0} & 74.98 & 80.96 & 100.0 & 92.91 & \textbf{91.45} & 90.05 & 9.29 & 9.69 \\
Xception Block 4 & 100.0 & 99.99 & 66.79 & 42.82 & 100.0 & \textbf{95.85} & 90.62 & 82.68 & 21.12 & 3.91 \\
Xception Block 5 & 100.0 & 100.0 & 60.44 & 48.92 & 100.0 & 93.09 & 89.08 & 81.92 & 19.92 & 4.11 \\
MesoInception4 & 100.0 & 97.90 & 49.72 & 45.98 & 98.71 & 80.57 & 71.27 & 74.02 & 20.84 & 3.55 \\
Resnet-18 & 100.0 & 64.80 & 47.06 & 54.69 & 79.20 & 51.15 & 52.37 & 58.21 & 10.84 & 5.37 \\
Xception & 100.0 & 99.75 & 55.85 & 40.98 & 99.94 & 85.69 & 74.33 & 76.09 & 21.86 & 3.48 \\ 
Xception++ & 99.95 & 91.65 & 81.47 & 85.89 & 94.00 & 75.54 & 64.18 & 84.67 & 11.24 & 7.53 \\
EfficientNet-V2 & 100.0 & 91.48 & 78.83 & 71.39 & 98.74 & 74.14 & 71.99 & 83.80 & 11.69 & 7.17 \\
CNN (p=0.1) & 100.0 & 98.41 & 90.46 & 50.65 & 99.95 & 90.48 & 85.27 & 85.87 & 16.53 & 5.20 \\
CNN (p=0.5) & 100.0 & 97.34 & \textbf{97.32} & 73.33 & 99.93 & 88.98 & 84.58 & \textbf{90.25} & 9.26 & 9.75 \\

\hline
\hline
\textbf{Ours} & 100.00 & 97.06 & 87.82 & 89.19 & 99.77 & 81.15 & 76.22 & 88.54 & \textbf{8.23} & \textbf{10.76} \\
\hline
\end{tabular}
\caption{\textbf{Synthetic Image Detection} Average precision on different model architectures and an alternative dataset (FFHQ). Excluding our results, all numbers taken from~\cite{chai2020makes}. As can be seen, our mean accuracy is good, and our consistency (low std) is the best. Combined, we achieve the best $CoV^{-1}$ score.}
\label{tab:SyntheticImagesAveragePrecisionWithCoV}

\end{center}
\end{table*}
\begin{table*}
            \footnotesize
                \begin{minipage}{0.6\textwidth}
                \raggedright
                
                \begin{tabular}{|l|c|c|c|c|c|c|c|c|c|c|c|c|c|c|}
\hline
\multicolumn{1}{|c|}{}& \multicolumn{4}{|c|}{\textbf{Train on Deepfakes}} & \multicolumn{4}{|c|}{\textbf{Train on Neural Textures}} \\

\textbf{Model} & \textbf{DF} & \textbf{NT} & \textbf{F2F} & \textbf{FS} & \textbf{DF} & \textbf{NT} & \textbf{F2F} & \textbf{FS} \\
\hline
\hline
Resnet Layer 1 & 98.97 & 74.99 & 71.74 & 57.15 & 70.32 & 86.93 & 65.04 & 52.3 \\
Xception Block 1 & 92.95 & 70.52 & 65.94 & 52.83& 66.30 & 80.72 & 62.65 & 52.0 \\
Xception Block 2 & 98.04 & 70.28 & 67.48 & 56.04 & 69.31 & 85.75 & 64.27 & 52.7 \\
Xception Block 3 & \textbf{99.41} & 67.58 & 63.62 & 57.97 & 67.62 & 85.44 & 60.71 & 52.0\\
Xception Block 4 & 99.14 & 68.91 & 70.36 & 58.74 & 73.65 & 90.97 & 60.72 & 52.7\\
Xception Block 5 & 99.27 & 68.25 & 66.68 & 43.20 & 83.52 & 92.23 & 63.75 & 49.9 \\
MesoInception4 & 97.28 & 59.27 & 60.17 & 47.24 & 65.75 & 83.27 & 62.92 & 54.0 \\
Resnet-18 & 93.90 & 53.22 & 53.45 & 53.69 & 69.98 & 85.40 & 54.77 & 50.8 \\
Xception & 98.60 & 60.15 & 56.84 & 46.12 & 70.07 & 93.61 & 56.79 & 48.5\\
Xception++ & 98.81 & \textbf{87.0} & \textbf{85.8} & 56.81 & 78.24 & 96.49 & 76.35 & 54.97 \\ 
CNN (p=0.1) & 97.78 & 60.08 & 59.73 & 50.87 & 68.67 & 95.16 & 68.15 & 47.4\\
CNN (p=0.5) & 98.16 & 54.02 & 56.06 & 55.99 & 66.98 & 95.03 & 71.50 & 51.9 \\
\hline
\hline
\textbf{\bf Ours} &
99.14 & \textbf{77.17} & 67.38 & \textbf{59.00} & \textbf{84.88} & \textbf{97.26} & \textbf{71.59} & \textbf{56.92} \\
\end{tabular}

\begin{tabular}{|l|c|c|c|c|c|c|c|c|c|c|c|}
\hline

\hline
\multicolumn{1}{|c|}{}& \multicolumn{4}{|c|}{\textbf{Train on Face2Face}} & \multicolumn{4}{|c|}{\textbf{Train on FaceSwap}} \\

\textbf{Model} & \textbf{DF} & \textbf{NT} & \textbf{F2F} & \textbf{FS} & \textbf{DF} & \textbf{NT} & \textbf{F2F} & \textbf{FS} \\
\hline
Resnet Layer 1 & \textbf{84.39} & 79.72 & 97.66 & 60.53 & 59.49 & 52.56 & 62.00 & 97.1 \\
Xception Block 1 & 77.65 & \textbf{80.88} & 93.84 & 61.62 & 53.14 & 49.24 & 56.89 & 82.8 \\
Xception Block 2 & 84.04 & 79.51 & 97.40 & 63.21 & 58.39 & 51.65 & 61.73 & 92.5 \\
Xception Block 3 & 76.10 & 74.77 & 97.33 & 63.10 & 61.77 & 53.44 & 61.34 & 96.0\\
Xception Block 4 & 67.18 & 61.72 & 97.19 & 63.04 & 61.33 & 52.02 & 59.45 & 96.5\\
Xception Block 5 & 81.25 & 61.91 & 96.45 & 55.15 & 57.14 & 47.39 & 54.68 & 95.5 \\
MesoInception4 & 67.53 & 55.17 & 92.27 & 54.06 & 50.64 & 48.87 & 56.15 & 93.8 \\
Resnet-18 & 55.43 & 52.57 & 93.27 & 53.39 &  61.03 & 51.66 & 52.56 & 91.4 \\
Xception & 66.12 & 56.07 & 97.41 & 53.15 & 53.86 & 50.00 & 56.55 & 96.8 \\
Xception++ & 68.77 & 71.65 & \textbf{98.92} & 67.03 & 48.03 & 53.06 & 69.15 & 99.47 \\ 
CNN (p=0.1) & 65.76 & 64.81 & 98.40 & 59.48 & 59.19 & 53.50 & 63.07 & 99.0 \\
CNN (p=0.5) & 65.43 & 60.36 & 97.94 & 63.52 & 60.19 & 52.11 & 59.81 & 98.2 \\
\hline
\hline
\textbf{\bf Ours} & 77.43 & 66.78 & 98.26 & 61.49 & \textbf{64.83} & \textbf{61.47} & \textbf{66.41} & \textbf{99.19} \\
\hline

\end{tabular}
                \end{minipage}
                ~~~~~~
                \begin{minipage}[c]{0.3\textwidth}

\begin{tabular}{|l|c|c|c|}
\hline

\hline
\multicolumn{1}{|c}{} &
\multicolumn{3}{|c|}{\textbf{Score}} \\

\textbf{Model} & Average & STD & $CoV^{-1}$ \\
\hline
Resnet Layer 1 & 73.18 & 15.54 & 4.71 \\
Xception Block 1 & 68.75 & \textbf{14.11} & 4.87 \\
Xception Block 2 & 72.02 & 15.07 & 4.78 \\
Xception Block 3 & 71.14 & 15.08 & 4.72 \\
Xception Block 4 & 70.85 & 15.57 & 4.55 \\
Xception Block 5 & 69.77 & 18.34 & 3.80 \\
MesoInception4 & 65.52 & 16.22 & 4.04\\
Resnet-18 & 64.16 & 16.20 & 3.96 \\
Xception & 66.29 & 18.43 & 3.60 \\
Xception++ & \textbf{75.66} & 16.87 & 4.48 \\
CNN (p=0.1) & 69.44 & 17.19 & 4.04 \\
CNN (p=0.5) & 69.20 & 17.04 & 4.06\\
\hline
\hline
\textbf{\bf Ours} & 75.57 & 14.94 & \textbf{5.06} \\
\hline

\end{tabular}
                \end{minipage}%
\caption{\textbf{Deepfake Detection} Average precision on FaceForensics++~\cite{roessler2019faceforensicspp} datasets when trained on one manipulation and tested against all four. (Left) results on all train/test combinations. (Right) Average, standard deviation (abbr. STD), and inverse CoV score of all models, given all combinations. Best scores are marked in bold. As can be seen, we achieve the second highest average accuracy ($75.57$) and the second lowest standard deviation ($14.94$), which leads to the highest overall $CoV^{-1}$ score ($5.06$). Interestingly, Xception with layer aggregation, denoted Xception++, performs better than vanilla Xception.}
\label{tab:DeepfakeTablesSidebySide}
\end{table*}

\subsection{Preprocessing} \label{SubSectionPreprocessing}
From each video, the first 100 frames are extracted. From each frame we cropped the face along with some context. Each cropped face was reshaped to $256 \times 256$ and aligned such that the left eye's center of mass is in the same location in all frames. 

For synthetic faces, we first generate the pristine and synthetic images as in~\cite{chai2020makes}. We follow a much simpler preprocessing procedure as we find the face using a simple face detector. The alignment is the same as in the Deepfake datasets: the left eye center of mass is always in the same location for all images.


\subsection{Model Architecture} \label{SubSectionModelArchitecture}
We use an EfficientNet-V2-Small~\cite{Tan2021EfficientNetV2SM} backbone. EfficientNet-V2-Small constitutes from a stack of MBConv~\cite{sandler2018mobilenetv2} and fused-MBConv~\cite{tan2021efficientnetv2} layers. The primitives extraction works as follows. From each MB-Conv layer we draw a feature vector. The feature vector is averaged with a 2D Average Pooling layer with a window of size $14$ for low level features and $7$ for high level features. Next, the pooled feature tensor is flattened and served as input to a three layered fully connected network (MLP). ReLU activations space each pair of fully connected layers. Since the stride changes in multiple stages of the EfficientNetV2-S, the height and width of the tensor changes. That means that each MLP has a different input dimension. The MLP output dimension, on the other hand, stays the same: $10$ for all features. Figure~\ref{fig:ModelArchitecture} (Right) depicts the primitives projection graphically.
Formally, to each layer $L_i$, $i=1, ..., L$ of the backbone we associate its output $o_i$ and input $o_{i-1}$, such that:
\begin{equation}
  o_i = L_i(o_{i-1})
\end{equation}
where $o_{0}$ is the input query image. For each output feature: $o_i$, $i = 1, ..., L$ we calculate its primitive projection. Formally, the primitives projections are defined $\forall i \in \{1, ...,L\}$:
\begin{equation}
\begin{aligned}
\vec{p}_i = & MLP(AvgPool2D(o_i))\\
\end{aligned}
\end{equation}
where $p_i \in \mathbb{R}^{10}$. The primitives are then aggregated to the final classification score with a linear regression block:
\begin{equation}
  s_{final} = w^T\cdot[\vec{p}_1, ..., \vec{p}_L] + b
\end{equation}
The network was trained with Binary Cross-Entropy loss:
\begin{equation}
\begin{aligned}
L = & CE(s_{final}, l) = \\
  & -[l \cdot log(s_{final}) + (1 - l) \cdot log(1 - s_{final}) ]
\end{aligned}
\end{equation}
Where $l\in \{0, 1\}$ is the label and $s_{final}$ is the binary score for the aggregated features.

\subsection{\bf Ranking results} 
The performance of an algorithm is usually measured using the Average Precision (AP) score. This works well for a single dataset but, in our case, we evaluate the same algorithm on multiple datasets. Therefore, we desire a measure of an algorithm across multiple datasets. To this end, we suggest the (inverse) Coefficient of Variation ($CoV^{-1}$). The $CoV^{-1}$ is defined as the average AP score, divided by its standard deviation. The numerator encourages accuracy (i.e., high AP score), while the denominator encourages consistency (low standard deviation).

\begin{figure}[t!]
\begin{center}
\includegraphics[height=0.25\textheight]{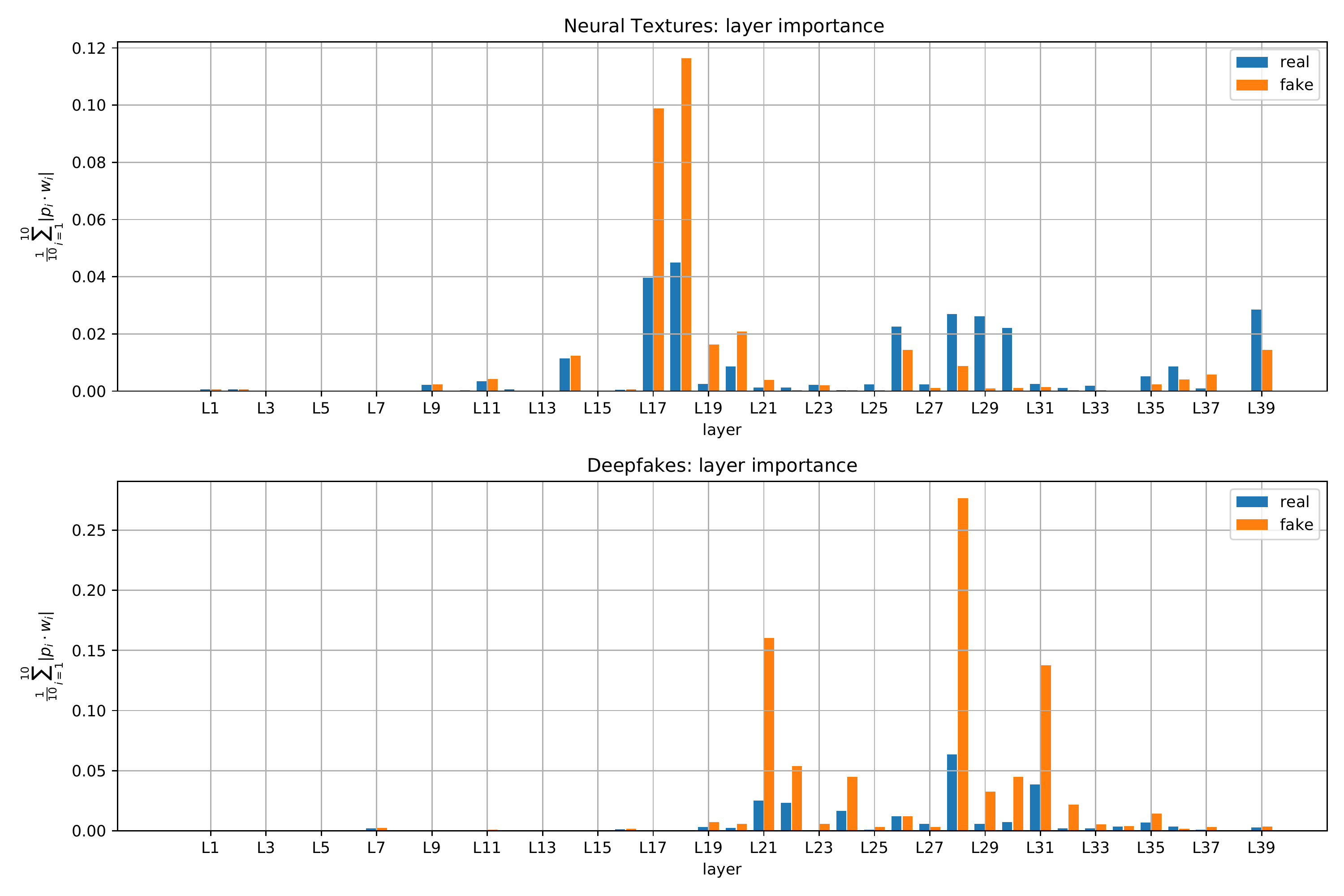}
\end{center}
\caption{\textbf{Deepfake Detection: Layer Importance}. Layers are indexed on the horizontal axis and importance (computed as the dot product of the weight of the Linear Regression and features of each layer) on the vertical axis. The average importance of $\sim1000$ real (fake) images is shown with blue (orange) bars. The graph on top (bottom) shows the average importance of a network trained with {\em NeuralTextures} ({\em Deepfakes}) models. As can be seen, the network 'automatically' calibrated itself to examine different features for different models. Namely, different features contribute to the detection of different manipulations.}
\label{fig:DeepfakesAnalysis}
\end{figure}

\section{Experiments}
\label{Experiments}
\subsection{Implementation Details}
We extract faces using a dlib-based face detector~\cite{dlibFaceDetector}. After that, we crop a bounding box around each face with a margin of $15\%$ from top and bottom, and $10\%$ from left and right. Then, each face is aligned using an affine warp such that the left eye is always in the same location and the image size is set to $256 \times 256$ (Code is available~\footnote{\url{https://github.com/ajev-research/agg-layers-deepfake-detection}}).

\subsection{Synthetic Image Detection}
We test the ability of different classifiers to generalize to different generator architectures. We follow~\cite{chai2020makes} to create a training set of PGAN fake images. We train our classifier based on the dataset of generated images and pristine images from CelebA-HQ. We test our generalization capability against dataset of real images from CelebA-HQ and synthetic images created from Style-GAN, Glow and GMM (all generating synthetic images with CelebA-HQ as a reference). Then, we also take natural images from FFHQ and build three more datasets of pristine and synthetic images using FFHQ and three generative models: PGAN, Style-GAN and Style-GAN2.

Table~\ref{tab:SyntheticImagesAveragePrecisionWithCoV} shows the Average Precision (AP) scores as reported in~\cite{chai2020makes} as well as the results of our architecture trained on PGAN and evaluated on the test set provided by~\cite{chai2020makes}. The rightmost columns of Table~\ref{tab:SyntheticImagesAveragePrecisionWithCoV} summarize the CoV scores. This single number helps us rank the synthetic image detection performance for each method with a single number. We out rank all models found in~\cite{chai2020makes}. In addition, we outperform a model created by taking EfficientNet-V2 and attaching a binary classification head on top of its $1000$ soft class prediction outputs ($10.76$ vs. $7.17$ $CoV^{-1}$ score). We let the latter train end-to-end. 

Using layer aggregation is not unique to EfficientNet architecture. We also report that adding layer aggregation to the Xception network improves the performance from $3.48$ to $7.53$. In Table~\ref{tab:SyntheticImagesAveragePrecisionWithCoV} we term it Xception++.

\subsection{Facial Manipulation}
We train our architecture on each of the four FaceForensics++ manipulations, and test generalization to the remaining three. Table~\ref{tab:DeepfakeTablesSidebySide} shows the Average Precision (AP) results when models are trained on one manipulation of FaceForensics++ and tested against all four manipulations. In total, we came first in 10 out of 16 tests, using the AP score. Our method achieves the second best average score (\ie~high accuracy) and the second lowest standard deviation (\ie~high consistency) leading to the best overall CoV score. See Table~\ref{tab:DeepfakeTablesSidebySide} (Right). Again, adding layer aggregation to Xception improves performance and the $CoV^{-1}$ score jumps from $3.60$ to $4.48$.

\subsection{Which features enable the detection of a fake?}
An interesting question to answer is which features enable the detection of a fake. Our network contains an inherent analysis tool to answer this question. We suggest to evaluate each features' performance from the product of its primitives $p_i$ with their corresponding weight of the Linear Regression weight vector, $w_i$. Our network is trained to output higher scores for fake images, and lower scores for real images. Therefore, $w_i\cdot p_i$ will be high for features that capture the "fakeness" property. The higher the energy of every such product the more meaningful it is for the classification.

Figure~\ref{fig:DeepfakesAnalysis} shows these products for Deepfakes and Neural Textures manipulations. These products are averaged over $\sim1000$ real images (shown in blue bars), and $\sim1000$ fake images (shown in orange bars). Observe, for example, that the features of layer $\#28$ are very important for real images in the Deepfakes model, while the features of layers $\#17-20$ play an important role in {\em NeuralTextures}. This shows that some primitives are more dominant than others, and usually when one primitive is active (that is, has high energy), then all primitives corresponding to its feature group are also active. This means that the feature was relevant for making the decision for the example image. Another interesting phenomenon is that for the two different manipulations, different primitives (and features, correspondingly) are active. Therefore, each manipulation has its own set of features which are useful to make the prediction.

\section{Applications}
\label{sec:Applications}
In addition to the ability to detect Deepfake images, we consider two additional applications. The first is analysis of the fake regions, and the other is a smart trimming approach of network size.

{\bf Fake region Analysis:} The method presented so far demonstrated how to automatically deduce which backbone-extracted-features are beneficial for the detection of fakes. Another important matter is to understand which regions in the input image contributed to the detection of the fakes. A common tool to answer this question is Class Activation Maps (CAMs)~\cite{zhou2016learning}. A class activation map is input specific and \emph{layer specific}. Our method unveils the layers that are important for classification thus the choice for the CAM layer is not arbitrary.

\begin{figure}
\begin{tabular}{ccc}
\includegraphics[height=0.11\textheight]{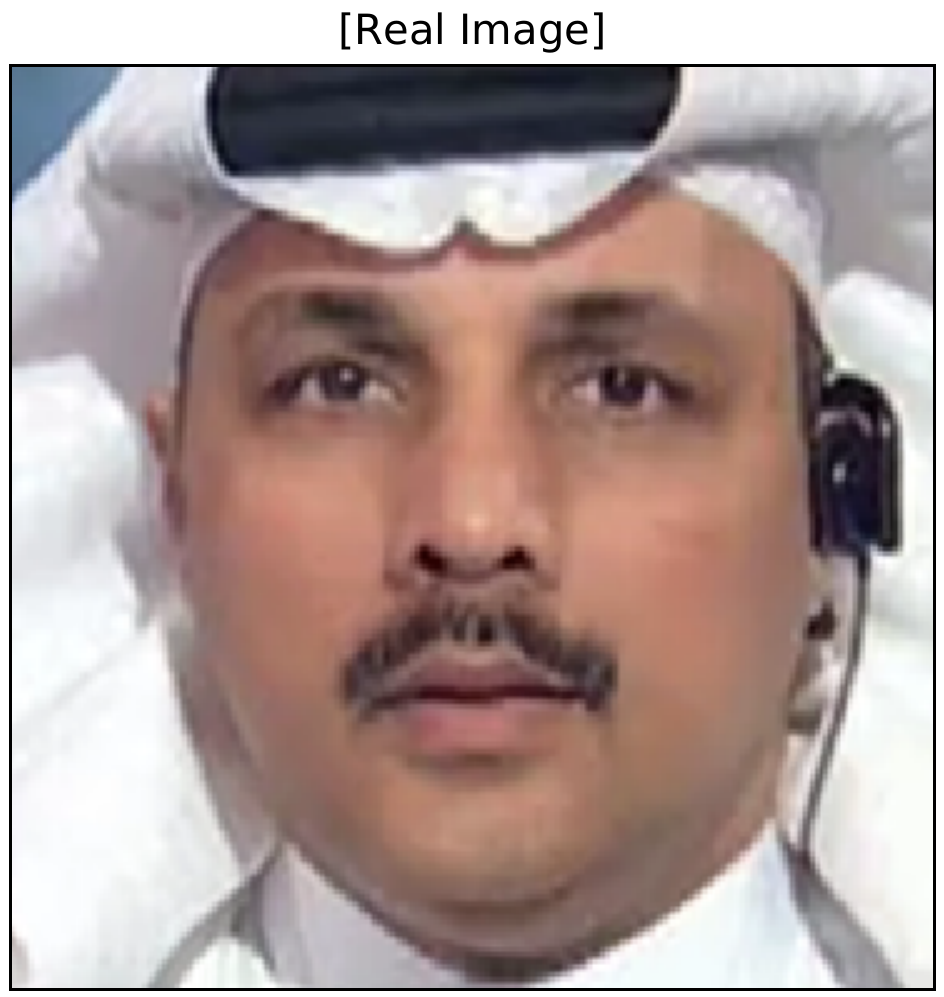} &
\includegraphics[height=0.11\textheight]{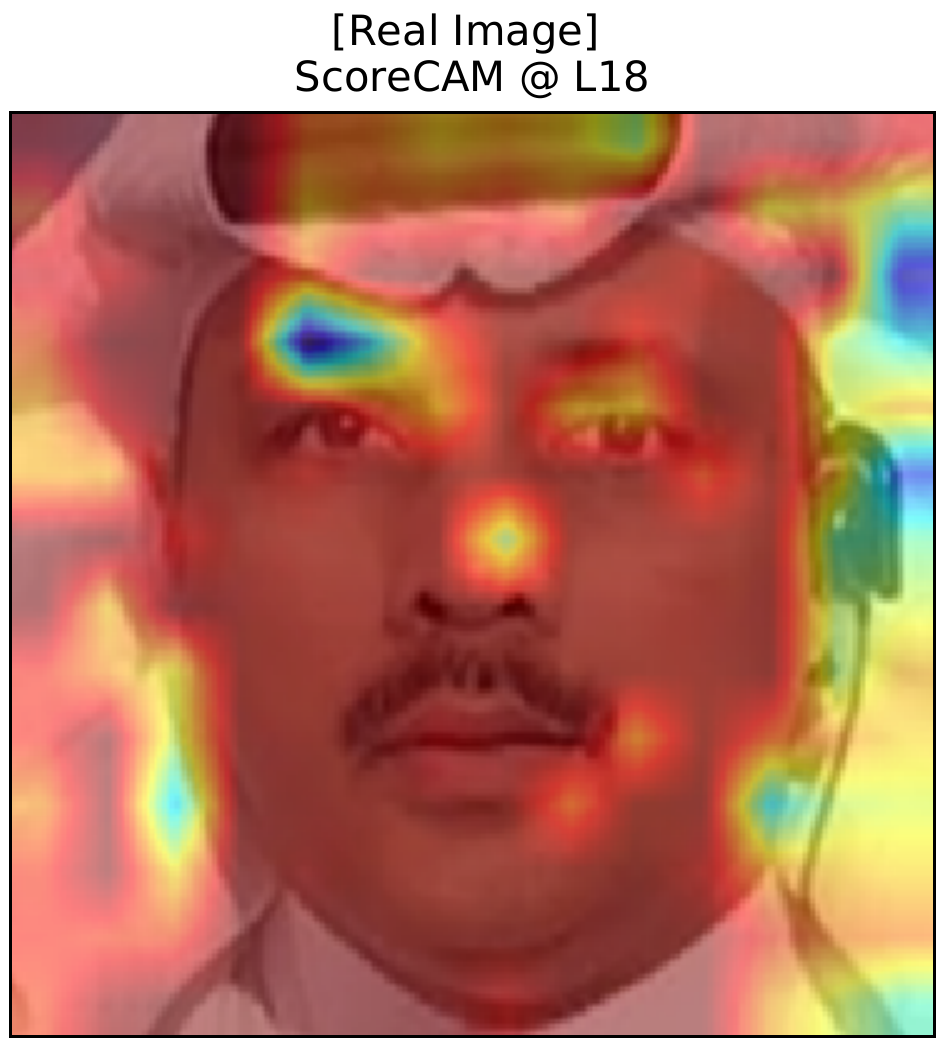} &
\includegraphics[height=0.11\textheight]{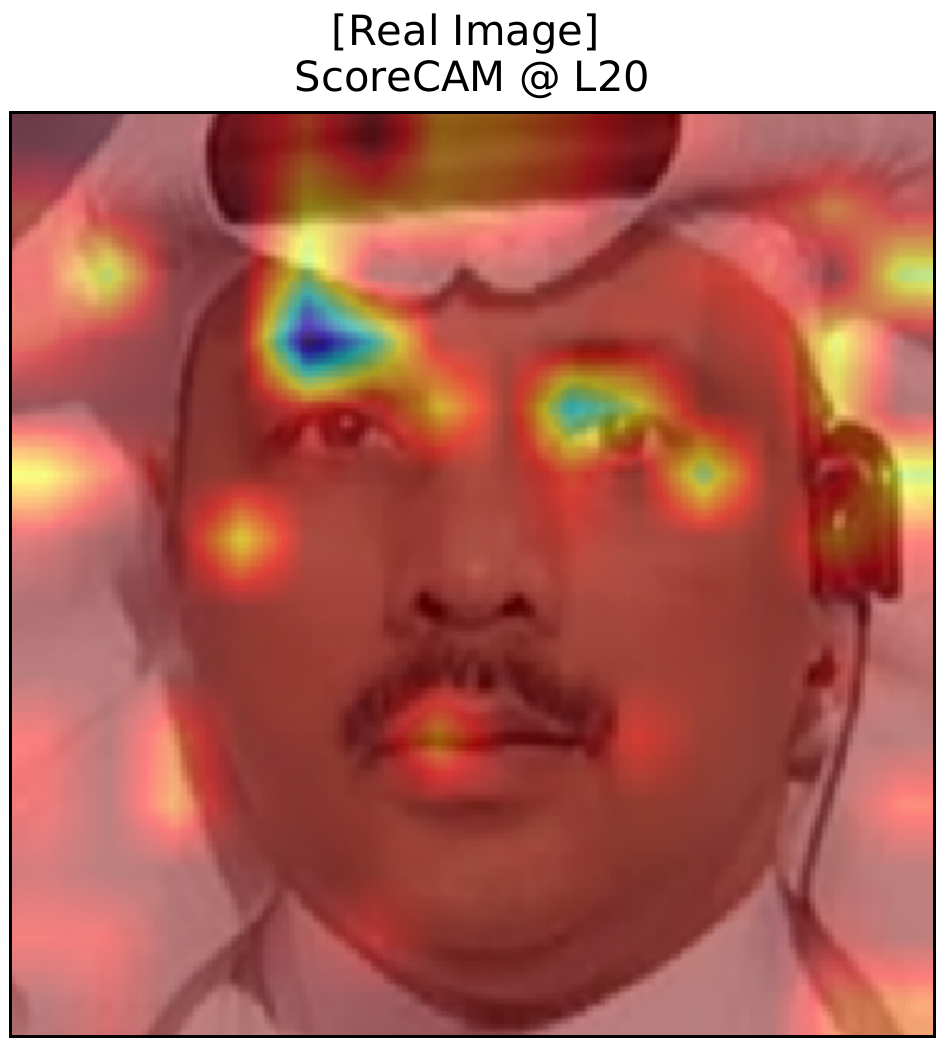} \\ 
\includegraphics[height=0.11\textheight]{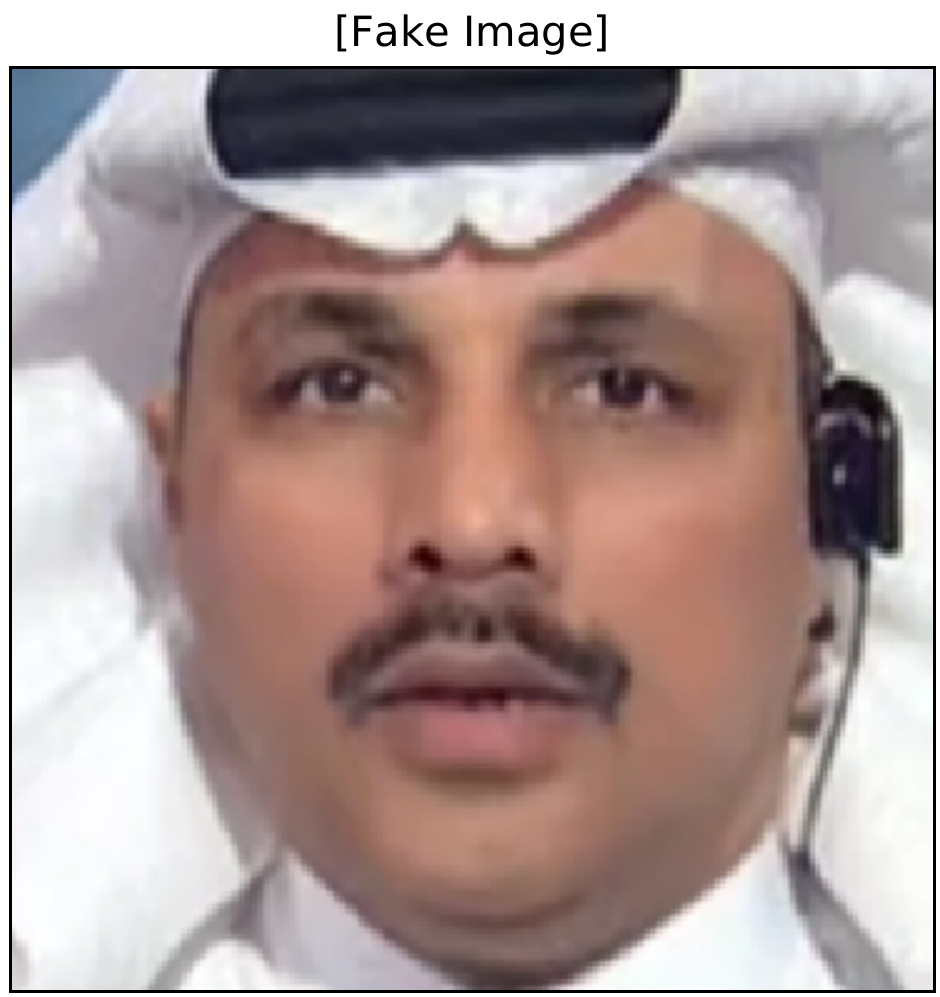} &  \includegraphics[height=0.11\textheight]{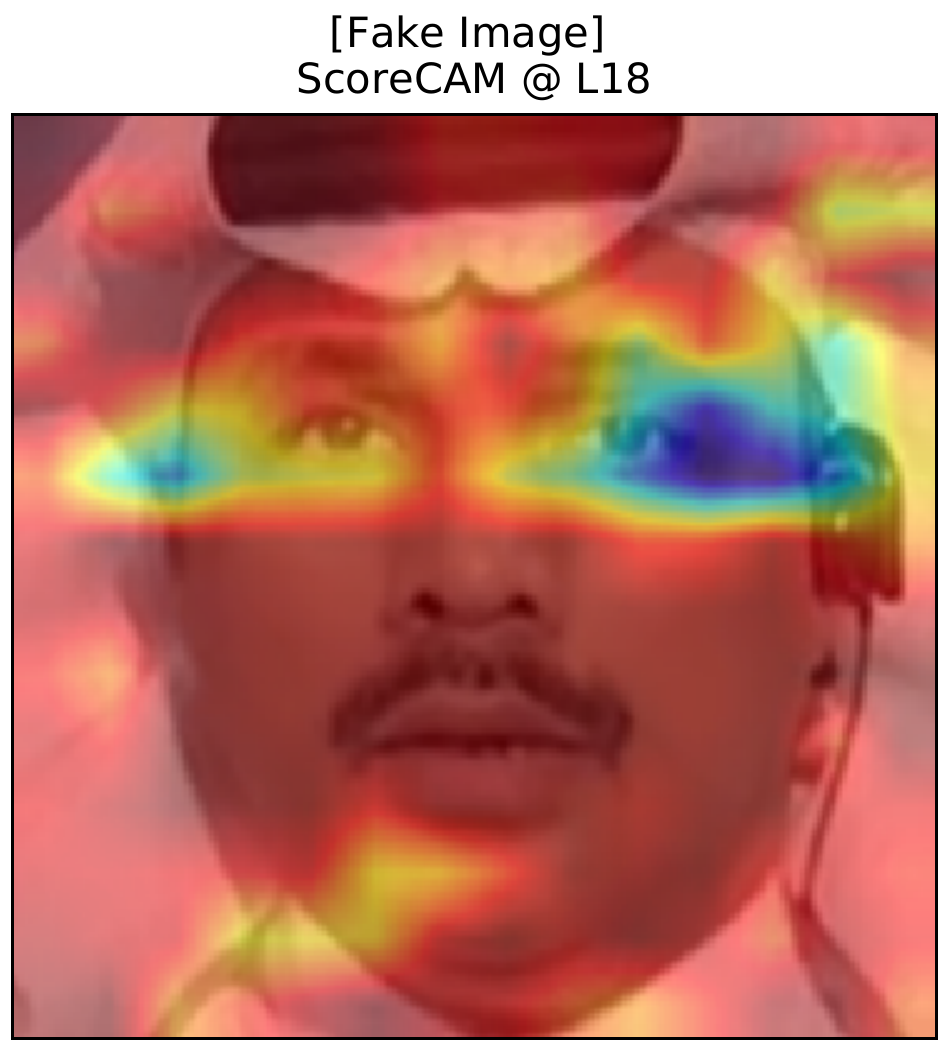} & \includegraphics[height=0.11\textheight]{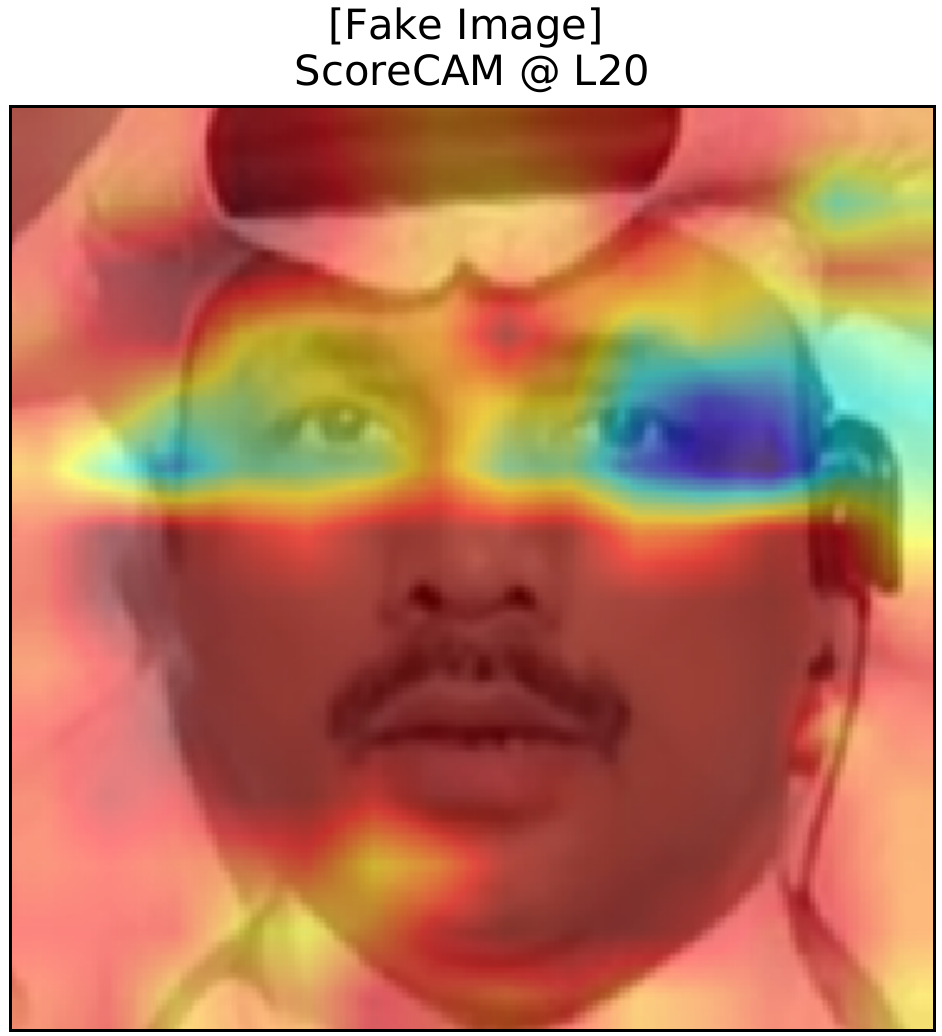}
\end{tabular}
\caption{\textbf{NeuralTextures: Class Activation Maps}. The top and bottom rows show the pristine and fake images, respectively. The fake image was generated with the Neural-Textures manipulation. The heatmaps in the second and third columns show Score-CAM~\cite{wang2020score} visualizations of layers 18 (middle) and 20 (right) in the network. The heatmaps show the importance of each pixel for classifying the image as fake. The eye region is detected as fake in this example.}
\label{tab:NeuralTexturesClassActivationMaps}
\end{figure}

\begin{figure}
\begin{tabular}{ccc}
\includegraphics[height=0.11\textheight]{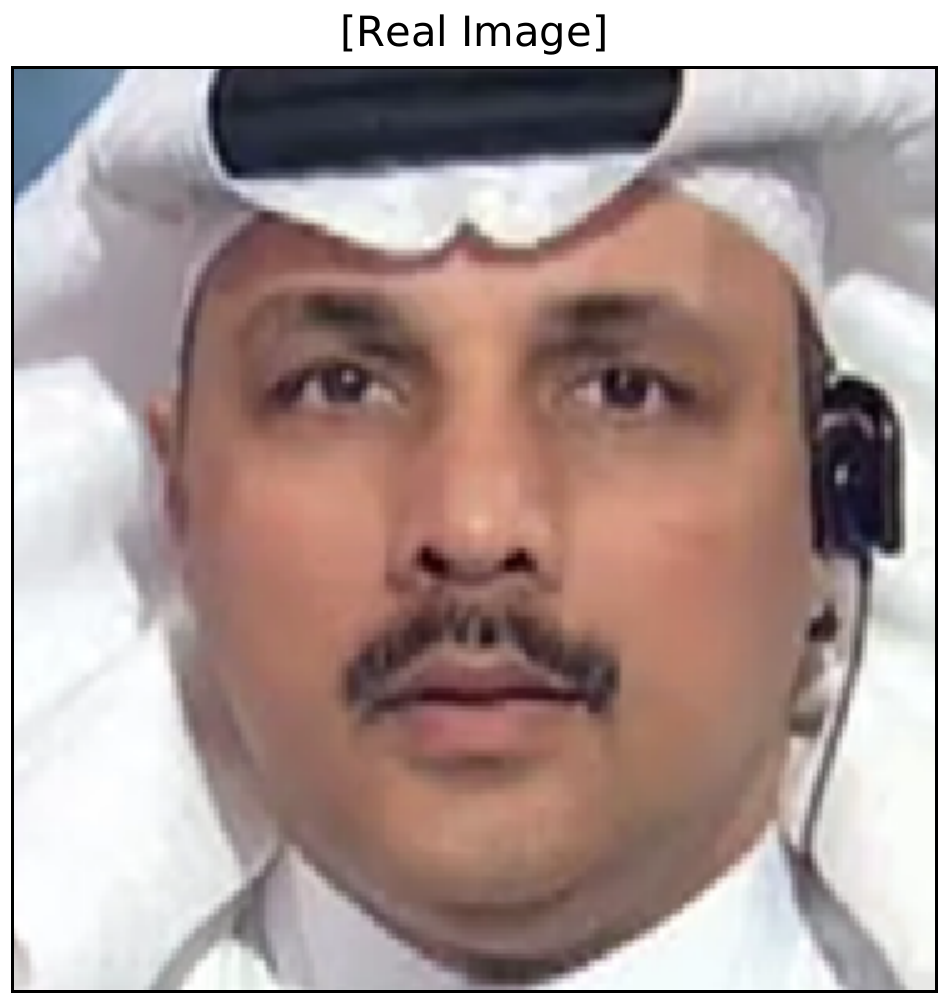} &
\includegraphics[height=0.11\textheight]{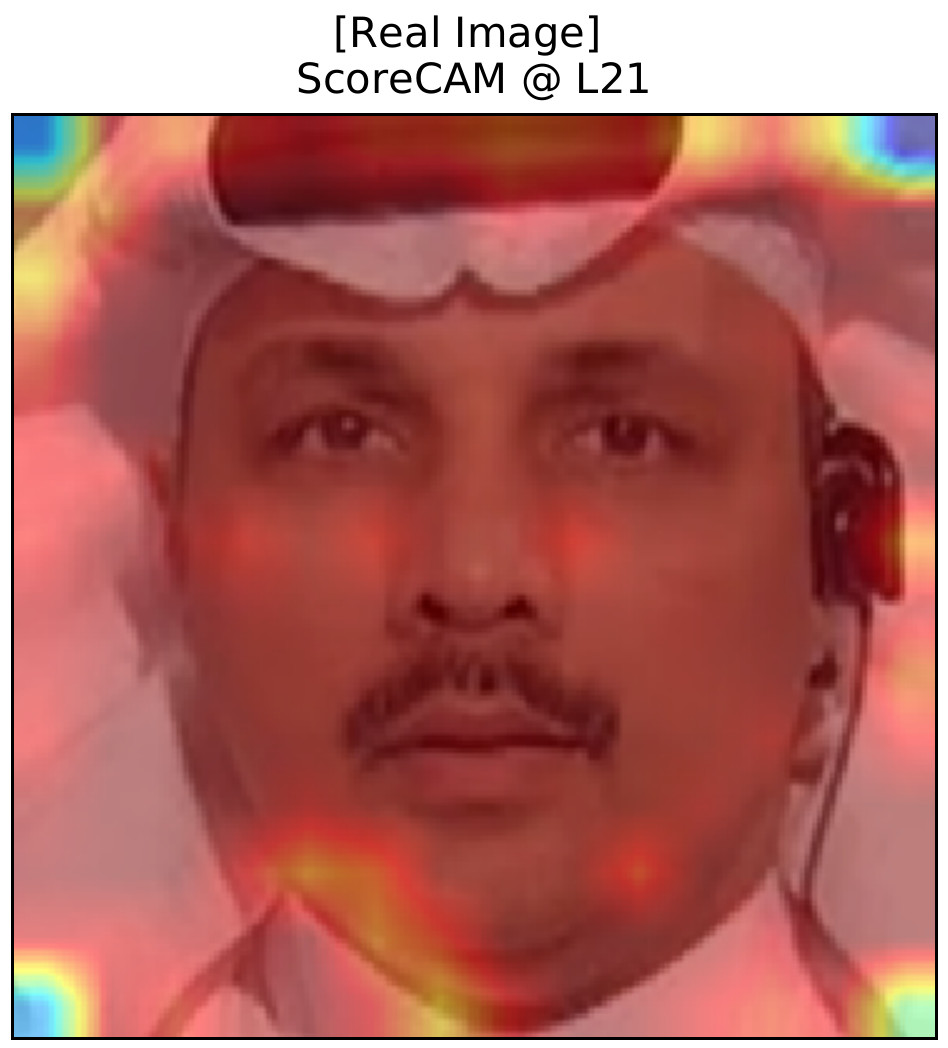} &
\includegraphics[height=0.11\textheight]{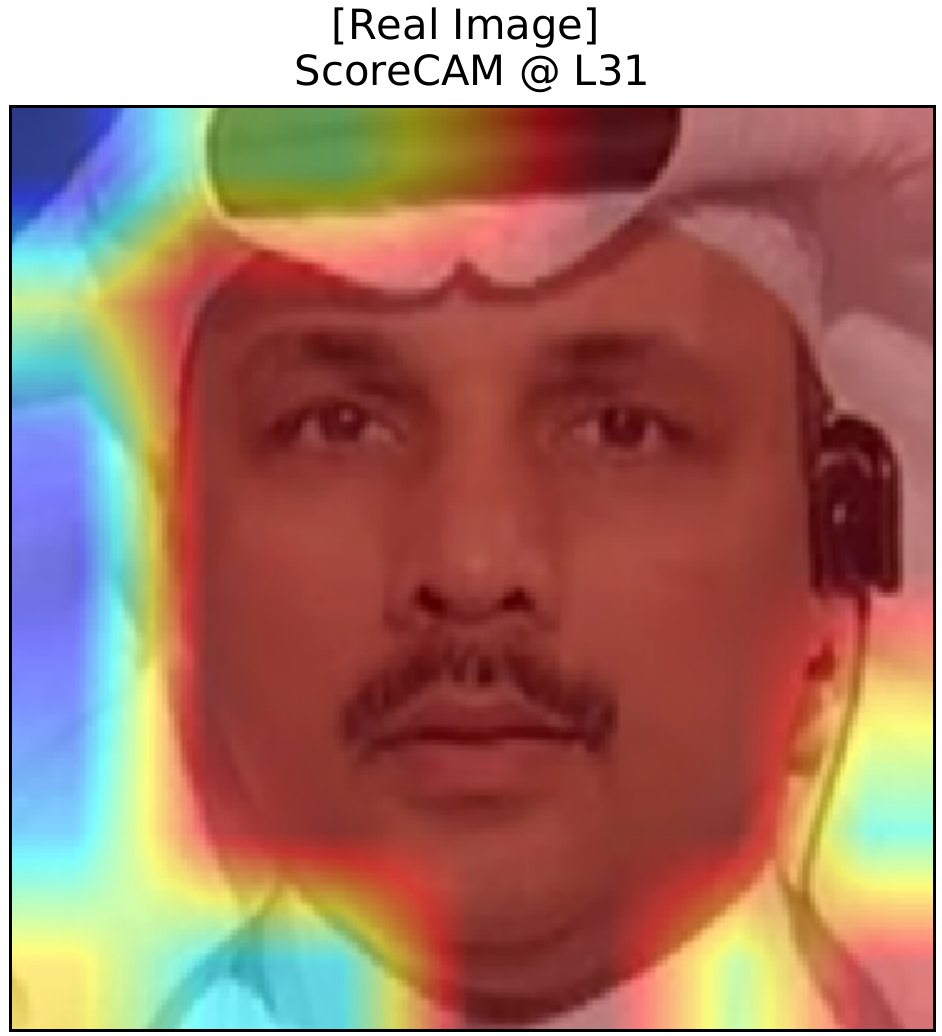} \\
\includegraphics[height=0.11\textheight]{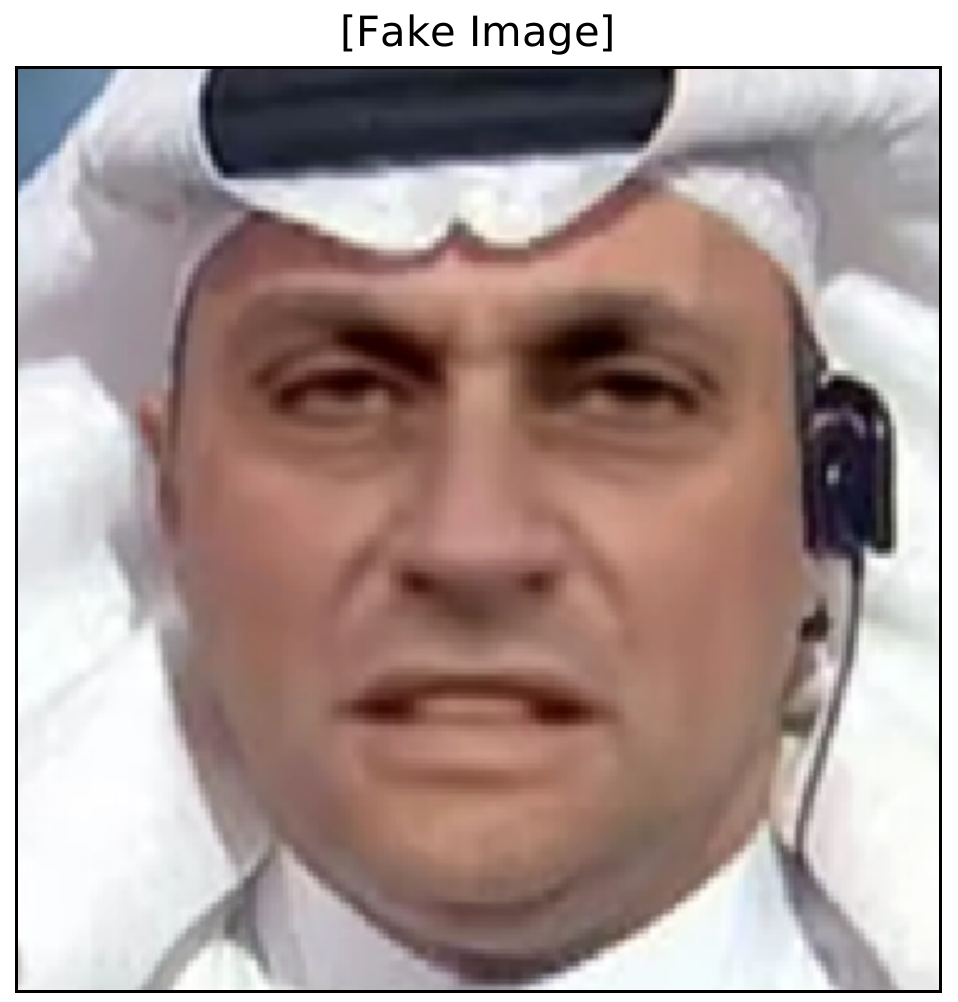} &
\includegraphics[height=0.11\textheight]{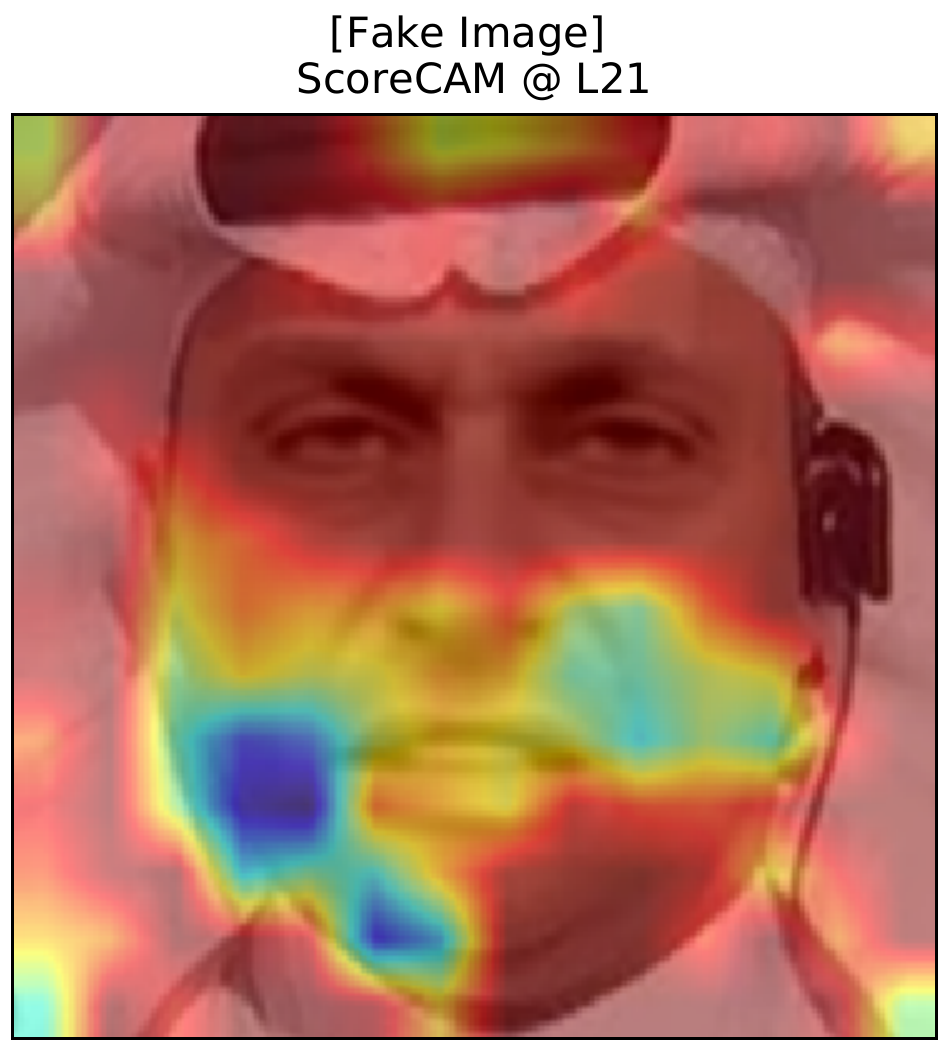} & \includegraphics[height=0.11\textheight]{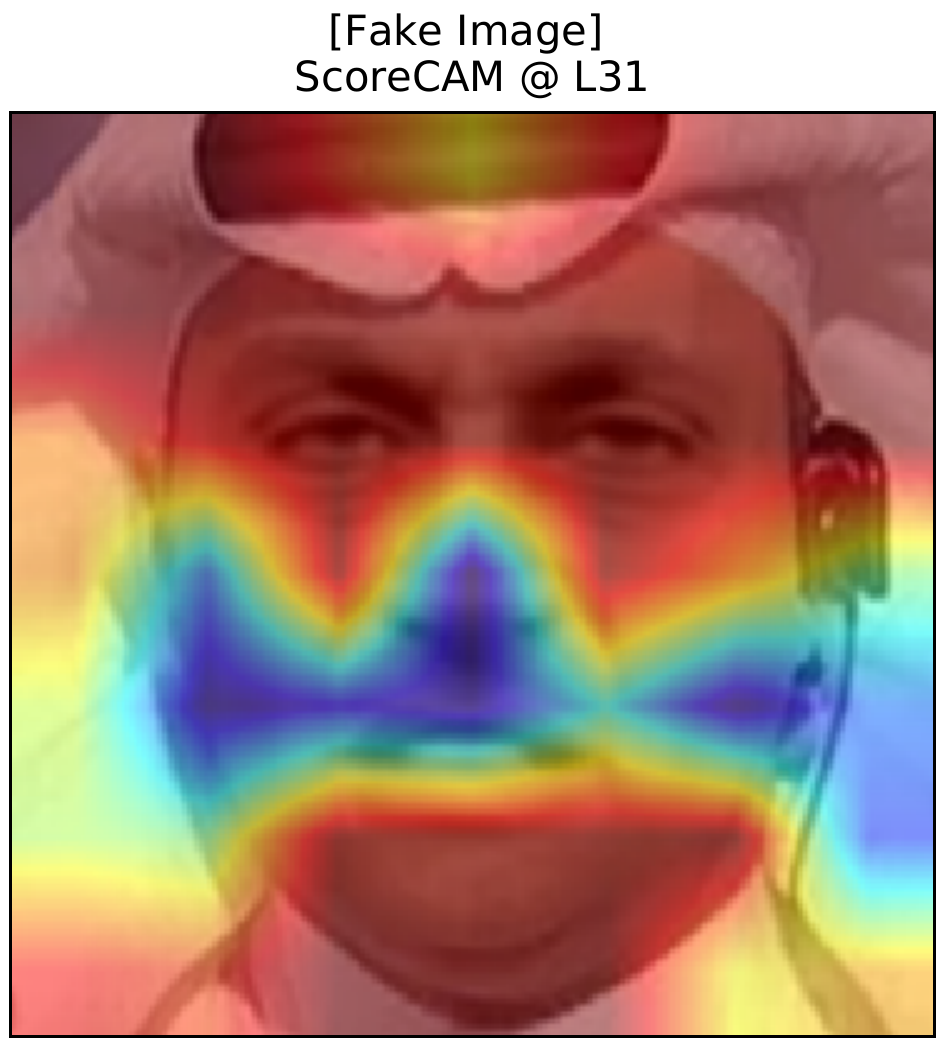}
\end{tabular}

\caption{\textbf{Deepfake: Class Activation Maps}. The top and bottom rows show the pristine and fake images, respectively. The fake generated with the Deepfake manipulation. The heatmaps in the second and third columns show Score-CAM~\cite{wang2020score} visualizations of layers 21 (middle) and 31 (right) in the network. The heatmaps show the importance of each pixel for classifying the image as fake. In this example, the fake region is around the mouth, nose and jaw.}
\label{tab:DeepfakesClassActivationMaps}
\end{figure}

We demonstrate this use case in Figures~\ref{tab:NeuralTexturesClassActivationMaps} and ~\ref{tab:DeepfakesClassActivationMaps} for {\em NeuralTextures} and {\em Deepfake}, respectively. In each case, we compute the most important feature layers for the input image and use CAM to generate an image-specific activation map. We observe that the eye region is detected as fake in the {\em NeuralTextures} example, while in the case of {\em Deepfake} the fake region is around the mouth, nose and jaw.


{\bf Network Trimming:} Our Deepfake network is based on EfficientNet-V2-Small as a backbone and an analysis component that consists of the projection primitive for each layer. The total number of parameters is about $40M$, where roughly $24M$ are for EfficientNet-V2-Small, and the remaining $16M$ parameters belong to the analysis part. Our goal is to trim the analysis part as much as possible.

The analysis part consists of 40 projection primitives, that correspond to the 40 layers of EfficientNet-V2-Small. Each projection primitive produces 10 features for a total of 400 features. Now, instead of keeping all 40 projection primitives, we only keep those whose features are important for correct deepfake classification.



\begin{figure}[htb]
\begin{center}
\includegraphics[height=0.14\textheight]{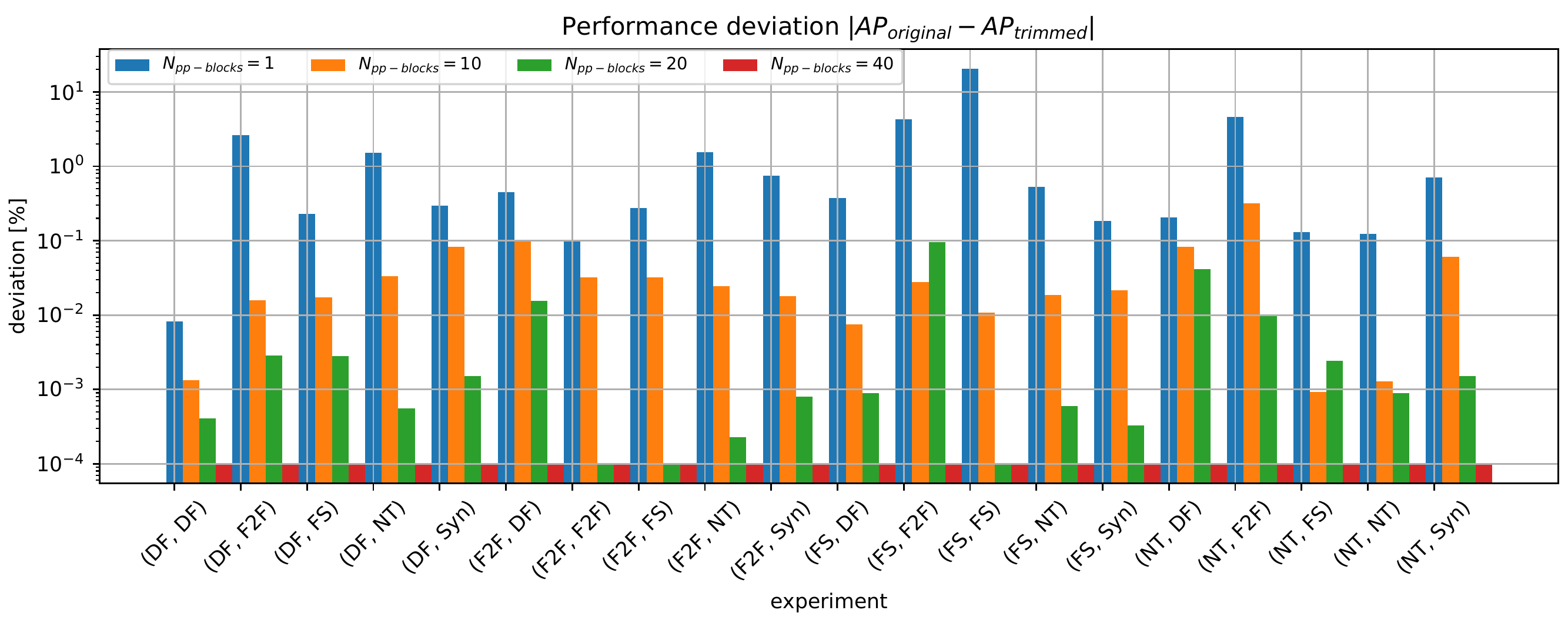}
\end{center}
\caption{\textbf{Deepfakes Network Trimming Effect}. Average Precision deviation vs. an experiment. The experiments are described as $(\mathcal{D} _{train}, \mathcal{D} _{test})$ pairs. The model is trained on $\mathcal{D} _{train}$ and tested on $\mathcal{D} _{test}$. The colors code different values of the Primitive-Projection blocks budget $N_{pp-blocks}$.}
\label{fig:PruningDegradationVsExperiment}
\end{figure}

We evaluated the Average Precision degradation by aggregating only the $N$ most important Primitive Projections. Figure~\ref{fig:PruningDegradationVsExperiment} shows the degradation of the trimmed analysis network w.r.t the full network for a various $N$s. Specifically, we report:
\begin{equation}
\begin{aligned}
AP_{deg} = 
\frac{1}{N_{exp}} \sum _ {(\mathcal{D} _{train}, \mathcal{D} _{test})}|AP_{entire} - AP_{trimmed}|
\end{aligned}
\end{equation}

Where $N_{exp}$ is the number of experiments, $AP_{entire}$ is the AP score as reported in the results section (the row marked {\bf "Ours"} in Table~\ref{tab:DeepfakeTablesSidebySide}(left)), $AP_{trimmed}$ is the AP score of the trimmed network and $\mathcal{D} _{train}$ and $ \mathcal{D} _{test}$ are the train and test datasets for the particular experiment. 
As can be seen, using just one Primitive Projection (the blue bars) yields the largest degradation across the different experiments, while using the best 20 Primitive Projections (the green bars) yields considerably smaller degradation.

\begin{figure}[htb]
\begin{center}
\includegraphics[height=0.27\textheight]{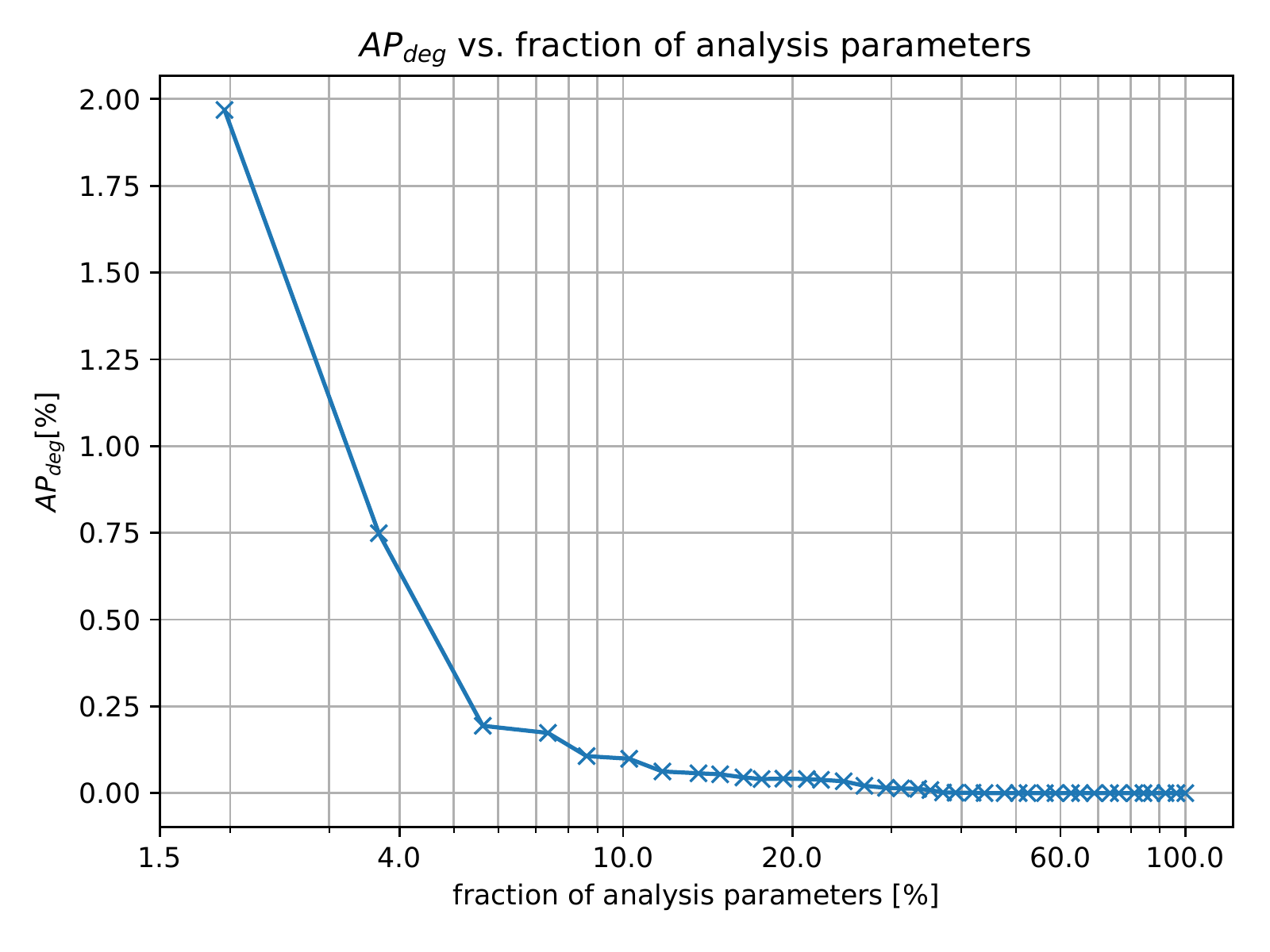}
\end{center}
\caption{\textbf{Deepfakes Trimming}. AP score degradation as a function of the number of analysis network parameters (The x-axis is presented in logarithmic scale). For example, choosing the best three Primitive Projection blocks requires less than $6\%$ of the total number of parameters in the analysis network and leads to a degradation of less than $0.25\%$, compared to using the entire analysis network of $16M$ parameters.}


\label{fig:DeepfakesPruningHeatmap}
\end{figure}

Next, we evaluate degradation in the AP score as a function of the trimming and report results in Figure~\ref{fig:DeepfakesPruningHeatmap}. The graph shows the degradation, as a function of the number of parameters of the analysis network. For example, using just the best Primitive Projection requires about $2\%$ of the analysis network (i.e., $0.32M$ parameters instead of $16M$) with a degradation in AP score of $2\%$. Using the top three Primitive Projection blocks requires less than $6\%$ of the analysis network (i.e., less than $1M$ parameters) with a degradation of less than $0.25\%$. We conclude that it is possible to achieve excellent classification results with a small analysis network of $1M$ parameters on top of a standard backbone such as EfficientNet-Small-V2.





\section{Conclusions}
Detecting synthetic and fake images attracts considerable amount of public interest. We consider the realistic scenario where a detection algorithm is trained on one Deepfake algorithm, and tested on fake images generated by other Deepfake algorithms. Similarly, the algorithm can be trained on synthetic images produced by one algorithm and detect synthetic images generated by another algorithm.

Our architecture is based on layer aggregation, a simple technique that aggregates information from all layers of a backbone network and feeds it to a classification head. As a by-product, our model lets us explore the features that cause the network to classify an image as real or fake. We then use this information to visualize the fake regions in the image plane. This lets the user understand which parts of the image where detected as fake. We also use the importance assigned per layer to trim layers that do not contribute to the Deepfake classification task.


We achieve SOTA results on multiple tests on multiple datasets, both for Deepfake detection and synthetic image detection. Results are summarized with a single number - the (inverse) Coefficient of Variance. This measure is defined as the ratio of average AP score (that encourages high AP score across all datasets) and low AP standard deviation (that encourages consistent results). Our method outperforms all other methods according to it.

\textbf{Acknowledgments:} This work was partly funded by ISF grant number 1549/19.

\bibliographystyle{ieee}
\bibliography{egbib}

\end{document}